\documentclass[twoside,dvipsnames]{article}

\usepackage[accepted]{aistats2024}

\usepackage[round]{natbib}
\renewcommand{\bibname}{References}
\renewcommand{\bibsection}{\subsubsection*{\bibname}}

\usepackage{hyperref}
\hypersetup{
	colorlinks,
	linkcolor={red!50!black},
	citecolor={blue!50!black},
	urlcolor={blue!80!black}
}

\usepackage{subcaption}
\usepackage[capitalise]{cleveref}
\usepackage{bbold}
\usepackage{mathtools}
\usepackage{pifont}%
\newcommand{\cmark}{{\color{tab10green}\ding{51}}\hspace{0.5em}}%
\newcommand{\xmark}{{\color{tab10red}\ding{55}}\hspace{0.5em}}%

\usepackage{booktabs}

\usepackage{multirow}
\usepackage{array}

\usepackage{xargs}
\usepackage[colorinlistoftodos,textsize=small]{todonotes} %
\newcommandx{\todoc}[2][1=]{{\todo[linecolor=orange,backgroundcolor=orange!25,bordercolor=orange,#1]{\small\textbf{TODO}: #2}}}
\newcommandx{\unsure}[2][1=]{{\todo[linecolor=yellow,backgroundcolor=yellow!25,bordercolor=yellow,#1]{\small\textbf{UNSURE}: #2}}}
\newcommandx{\change}[2][1=]{{\todo[linecolor=blue,backgroundcolor=blue!25,bordercolor=blue,#1]{\small\textbf{CHANGE}: #2}}}
\newcommandx{\info}[2][1=]{{\todo[linecolor=green,backgroundcolor=green!25,bordercolor=green,#1]{\small\textbf{INFO}: #2}}}
\newcommandx{\improvement}[2][1=]{{\todo[linecolor=violet,backgroundcolor=violet!25,bordercolor=violet,#1]{\small\textbf{IMPROVEMENT}: #2}}}
\newcommandx{\thiswillnotshow}[2][1=]{{\small\todo[disable,#1]{THIS WILL NOT SHOW: #2}}}

\usepackage{algorithm} \usepackage[noend]{algpseudocode}

\definecolor{rowhighlightcolor}{HTML}{ffecb3}

\usepackage{wrapfig}

\usepackage{bbm}

\usepackage{annotate-equations}

\usetikzlibrary{calc}

\newcommand{\data}{\mathcal{D}}
\newcommand{\datasynth}{\widetilde{\mathcal{D}}}

\newcommand{\teachermodel}{f^{\text{T}}}
\newcommand{\studentmodel}{f^{\text{S}}}

\newcommand{\teacherlogits}{\mbz^\text{T}}
\newcommand{\studentlogits}{\mbz^\text{S}}
\newcommand{\xsynth}{\widetilde{\mbx}}
\newcommand{\ysynth}{\widetilde{y}}
\newcommand{\fn}[2]{#1\!\left( #2 \right)}

\newcommand{\mathbold}[1]{\boldsymbol{#1}}

\newcommand{\mbx}{\mathbold{x}}

\newcommand{\mbz}{\mathbold{z}}

\definecolor{tab10green}{HTML}{2CA02C}
\definecolor{tab10blue}{HTML}{1f77b4}
\definecolor{tab10red}{HTML}{d62728}

\definecolor{teacher-cmap-green}{rgb}{0.2923363702661794,0.5107465581015513, 0.3906173846714218}
\definecolor{teacher-cmap-purple}{rgb}{0.6613156436870031, 0.33633419891183414, 0.7164828207890344}
\definecolor{student-cmap-blue}{rgb}{0.24715576253545807, 0.49918708160096675, 0.5765599057376697}
\definecolor{student-cmap-red}{rgb}{0.7634747047461135, 0.3348456555528834, 0.225892295531744}

\usepackage{float}  %
\usepackage{lipsum}

\begin{document}

\runningauthor{Steven Braun, Martin Mundt, Kristian Kersting}

\twocolumn[

	\aistatstitle{Deep Classifier Mimicry without Data Access}

	\aistatsauthor{ Steven Braun\textsuperscript{\normalfont 1} \And Martin Mundt\textsuperscript{\normalfont 1,2} \And  Kristian Kersting\textsuperscript{\normalfont 1,2,3,4} }

	\aistatsaddress{
\\
	\textsuperscript{1}Department of Computer Science, TU Darmstadt, Darmstadt, Germany\\
  \textsuperscript{2}Hessian Center for AI (hessian.AI), Darmstadt, Germany\\
  \textsuperscript{3}German Research Center for Artificial Intelligence (DFKI), Darmstadt, Germany\\
  \textsuperscript{4}Centre for Cognitive Science, TU Darmstadt, Darmstadt,
	Germany\\
	  \texttt{\{steven.braun,martin.mundt,kersting\}@cs.tu-darmstadt.de}
 }
	]

\begin{abstract}
	Access to pre-trained models has recently emerged as a standard across numerous
	machine learning domains. Unfortunately, access to the
	original data the models were trained on may not equally be granted.
	This makes it tremendously challenging to fine-tune, compress models,
	adapt continually, or to do any other type of data-driven update. We posit
	that original data access may however not be required. Specifically, we propose Contrastive
	Abductive Knowledge Extraction (CAKE), a model-agnostic knowledge
	distillation procedure that mimics deep classifiers without access to the
	original data. To this end, CAKE generates pairs of noisy synthetic samples and
	diffuses them contrastively toward a model’s decision boundary. We
	empirically corroborate CAKE's effectiveness using several benchmark
	datasets and various architectural choices, paving the way for broad
	application.
\end{abstract}

\section{INTRODUCTION}
\label{sec:introduction}

In the contemporary machine learning landscape, the rise in availability of pre-trained models
has significantly facilitated development of downstream applications. In conjunction with prominent underlying techniques, ranging from parameter pruning and sharing~\citep{han2015learn,wang2016cnnpack}, low-rank factorization~\citep{yu2017compressing,denton2014exploiting}, to knowledge
distillation~\citep{hinton2015distill}, these pre-trained models can now be efficiently fine-tuned, compressed, or even adapted continually. Enabling all the latter through a single mechanism, knowledge distillation seems to be a particularly promising contender from the plethora of available options. At its core, it aims to transfer the knowledge from a (typically larger, more complex) teacher model to a (typically smaller, simpler) student model by training the student to mimic the teacher's predictions, feature responses, or other inferrable quantities from the learned function. Such mimicry then enables the student to reach similar performance levels, at reduced computational cost and memory usage or allow a model to continue learning, if the student is the same model that retains knowledge from a prior time step~\citep{Li2016LearningWF}. However, the knowledge distillation optimization procedure traditionally requires access to original data. Unfortunately, a provided model state may not be accompanied with all its training data or access to the latter may deliberately not be granted.

Despite an impressive amount of ensuing applications in natural language processing~\citep{tang2019distilling-bert,mou2016word-emb,jiao2019tiny-bert},
computer vision~\citep{liu2017learning-efficient-cnn,zhou2018dorefanet-tl,yim2017agf},
and speech recognition~\citep{hinton2015distill,lu2016highway,ramsay2018low-dimensional-bf}, the majority of approaches is thus still limited by a host of assumptions. In most works, students train on the original training data~\citep{hinton2015distill,chen2019dafl,han2020robustness-ad} or additional generative auxiliary models are used to approximate the data distribution~\citep{chen2019dafl,han2020robustness-ad}. Alternatively, the necessity of data can be alleviated by imposing heavy constraints on architectures~\citep{chen2019dafl,han2020robustness-ad,hongxu2019deepinversion,nayak19azskd}. These dependencies limit the applicability of knowledge distillation when the original training
data is not available, the teacher and student model architectures mismatch, or
training additional generative models is infeasible. However, a crucial realization is that a majority of these tasks may not even require strong assumptions if one accounts for the task's nature.

Specifically, \emph{we posit that supervised scenarios and classification, in particular, do not require all knowledge to be distilled}. On the contrary, it is \emph{the decision boundary that needs to be closely mimicked} by a student. We refer to respective distillation as \textbf{abductive knowledge extraction}.

\begin{figure*}
	\centering
	\input{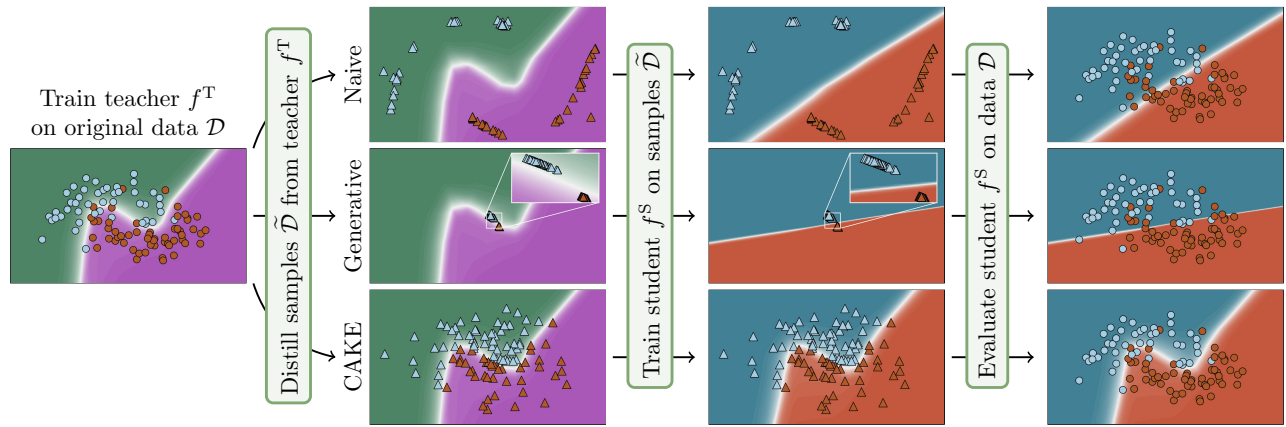}
	\caption{Comparison of naive, generative, and CAKE methods for knowledge distillation on the two-moons dataset. The background visualizes teacher ({\color{teacher-cmap-green}green}/{\color{teacher-cmap-purple}purple}) and student ({\color{student-cmap-blue}blue}/{\color{student-cmap-red}red}) decision functions, juxtaposed with original data ($\circ$) and synthesized samples ($\triangle$). Naive and generative methods often converge to similar local minima, inducing an ineffective student decision function. In contrast, CAKE generates samples across the entire decision-relevant region, resulting in a student model that accurately learns the data decision function if trained exclusively on its synthetic samples.}
	\label{fig:synth-examples}
\end{figure*}

Based on this realization, we lift prior works' assumptions and propose the
first knowledge distillation method that is truly model-agnostic, i.e. operating
effectively without reliance on any specific architectural features, components,
or configurations of the teacher (or the student) model, while not requiring access to any original training data.
To this end, our introduced Contrastive Abductive Knowledge Extraction (CAKE) generates synthetic data pairs via a contrastive diffusion process, which are directed toward opposing sides of a teacher model's decision boundary. In symbiosis, a contrastive pull ensures that a prospective student trains on samples that closely support the teacher's decision, whereas induced noise scatters samples to sweep relevant regions along the boundary. \cref{fig:synth-examples} shows an intuitive ``two-moons'' example, where CAKE is compared to naive synthetic samples based on gradient descent alone and a generative model. As detailed later, CAKE succeeds where competitors fail at data-free model-agnostic knowledge distillation due to collapse to trivial solutions or failure to cover a broad spectrum close enough to the relevant decision boundary.

\begin{itemize}
	\item We introduce Contrastive Abductive Knowledge Extraction (CAKE), a model-agnostic knowledge distillation procedure without access to original data. Instead, a contrastive diffusion process generates synthetic samples that border a teacher's decision boundary.
	\item We empirically highlight the contribution of CAKE's components, showcase how teacher and student neural networks can differ in depth and capacity, and analyze CAKE's effectiveness when teacher and student models differ (MLP, CNN, ResNet, and ViT).
	\item We corroborate that CAKE's classification accuracy is competitive with a variety of ``state-of-the-art'' methods that require data access or heavy model assumptions.
\end{itemize}

We provide open-source code for CAKE at \url{https://github.com/ml-research/CAKE}.

\section{KNOWLEDGE DISTILLATION AND THE CHALLENGE OF DATA AVAILABILITY}
\label{sec:knowledge-distillation}

In this section, we will discuss the key concepts behind knowledge distillation,
briefly explore the different types of distilled knowledge and distillation
schemes, and summarize limitations with respect to data availability commonly
found in the literature and respective surveys~\citep{gou2020kd-survey,liu2021datafree}.

\subsection{Distillation in Supervised Classification}
The original variant of knowledge distillation introduced by
\citet{hinton2015distill} uses a softened version of the teacher's (logit) output
to train a student model to mimic the teacher. At the example of supervised classification, given a training dataset with $N$ input-target pairs
$(\mbx_i, y_i)$, a student $\studentmodel$, and a teacher $\teachermodel$, we denote
$\studentlogits_{i} = \studentmodel\!\left( \mbx_{i}\right)$ and
$\teacherlogits_{i} = \teachermodel\!\left( \mbx_{i}\right)$ as the student and teacher
logits respectively. The student is trained by minimizing a loss
function $\mathcal{L}$
that balances the prediction of ground truth labels and matches the softened
output of the teacher $\fn{p}{\mbz_{i}, \tau } = \left( \exp\left(z^{1}_{i}/\tau\right)  / Z_{i}, \dots, \exp\left(z^{C}_{i}/\tau\right)/ Z_{i} \right)$,
where $Z_{i} = \sum_{j} \exp(z^{j}_{i}/\tau)$ is the normalization constant and $\tau$ is a temperature parameter that controls the softness of the output
distribution. The full student training objective thus becomes a conjunction of true labels and the teacher's ``soft labels'':
\begin{multline}
	\label{eq:hinton-loss}
	\mathcal{L}\left(\mbx_i, y_i\right) =
	\lambda_{1}
	\fn{\text{CE}}{y_i , \fn{p}{\studentlogits_{i} , 1 } } \\
	+ \lambda_{2}
	\fn{\text{CE}}{\fn{p}{\teacherlogits_{i} , \tau}, \fn{p}{\studentlogits_{i} , \tau}} \quad ,
\end{multline}
where $\lambda_{1}$ and $\lambda_{2}$ are hyperparameters that control the
trade-off between the two terms and CE is the cross-entropy loss. The
first term ($\mathcal{L}_{\text{NLL}}$) in the loss function encourages the student to predict the
ground truth labels, while the second term ($\mathcal{L}_{\text{KD}}$) tries to
match the softened output of the teacher.

More generally, knowledge distillation techniques can be categorized based on the \emph{distilled
	knowledge} and \emph{distillation schemes}. Distilled knowledge may include response-based
methods focusing on model outputs~\citep{hinton2015distill}, feature-based
methods targeting intermediate
representations~\citep{zagoruyko2017paying,chen2017object-det-kd}, and
relation-based methods capturing pairwise relations between
instances~\citep{yim2017agf,you2017multiple-teacher}. Distillation schemes may
encompass offline distillation, which involves pre-trained teacher models,
online distillation where teacher and student models are trained simultaneously,
and self-distillation, where the teacher and student are the same
model~\citep{zhang2019self-distillation,hou2019self-attention-distillation,yang2018snapshot-distillation}.
The choice of the distillation scheme depends on an application's
requirements, including computational resources, training data, and desired
accuracy.

\subsection{Distillation without Data Access}
\label{sec:knowledge-distillation:data-free}

While the seminal work of \citet{hinton2015distill} introduced knowledge
distillation with the student having access to the original training data and
using the smoothed teacher outputs as \emph{additional information}, knowledge
distillation can be further lifted to a \emph{``data-free''} setting. Here,
data-free refers to providing no access to the data distribution that the
teacher was trained on. The focus is then to construct synthetic samples from
the teacher that serve as exclusive training data for the student. There are
generally two approaches to achieve such generation of synthetic data.

One angle makes use of generative models to synthesize samples that are relevant
to the teacher's objective, therefore extracting knowledge into an auxiliary
generative model that learns to sample the data distribution. Recent examples
are adversarial distillation, such as DAFL~\citep{chen2019dafl} and its
successor RDSKD~\citep{han2020robustness-ad} which employ a generative
adversarial network (GAN)~\cite{Goodfellow2014GenerativeAN}. However, while
students are trained with synthetic GAN samples, the training procedure of the
GAN itself again requires access to original data to construct the adversarial
objective. Further, an additional model now needs to be carefully crafted, which
may be prone to issues such as e.g. mode collapse in GANs.

The alternative angle is to leverage the teacher's parameters directly to
construct a synthetic dataset. The initial work,
DeepDream~\citep{mordvintsev2015deepdream}, uses an image prior and performs
gradient descent on an input image w.r.t. maximizing a specific class
probability. The later DeepInversion~\citep{hongxu2019deepinversion} uses
DeepDream's total variation and the $l_{2}$-norm as an image prior and extends
the optimization objective by a feature distribution regularization term. This
imperative term measures the $l_{2}$-distance between convolution activations
and respective BatchNorm~\citep{ioffe2015batchnorm} statistics, as the latter
provides a simple Gaussian proxy to the encoded distribution.
\cite{Wang2021DataFreeKD} extends this principle by modeling the intermediate
feature space with a multivariate normal distribution. However, methods such as
DeepInversion then entail a restriction on the teacher requiring specific
layers. As such, the approach cannot be applied if the teacher is either treated
as a black box with no access to intermediate outputs or does not contain the
necessary functions. CAKE follows in these works' footsteps, but lifts the
constraints on model architecture and intermediate value access.

\subsection{The Pitfalls when Removing Data Access without Model Constraints}
To contextualize prior works and highlight the challenge of removing both access
to data and intermediate values of specific model functions, we circle back to
the earlier shown figure \cref{fig:synth-examples}. On the basis of the simple
2-D two-moons example, the top row depicts original (circles) and synthetic
(triangles) data for the naive DeepDream approach. As the latter optimizes
initially random inputs solely to maximize the cross-entropy loss, the first
common pitfall ensues. Namely, \emph{samples easily satisfy maximum confidence
	if they lie far away from the decision boundary}. When a student is trained on
these samples, the decision boundary is overly simplistic, here leading to a
linear decision that is incorrect for the original task. The second row shows a
respective generative model trained to synthesize data that minimize the
teacher's confidence. Whereas the first pitfall may also occur, we can condition
samples and contrast pairs (as in the later CAKE for direct comparison).
However, a second caveat now arises. Namely, \emph{parameterized generators may
	easily collapse towards trivial solutions or sample select regions}. As they
collapse to specific modes that do not cover the distribution necessarily, the
student's solution may once more be inadequate for the original task.

\section{CAKE: CONTRASTIVE ABDUCTIVE KNOWLEDGE EXTRACTION}
\label{sec:cake}

In the previous section, we expounded on the limitations and
assumptions associated with existing knowledge distillation techniques when
original training data is unavailable and strict model assumptions cannot be made.
To overcome these challenges, we now introduce Contrastive Abductive Knowledge Extraction, CAKE for short. In contrast to prior works, CAKE extracts the abductive knowledge of a teacher in a fully model-agnostic way that does not require any original data.

\subsection{Contrasting the Decision Boundary: Abductive Knowledge Extraction}
\label{sec:cake:extraction-objectives}

We propose a conceptual shift in the objective of the distillation procedure.
Contrary to the emphasis placed by a significant portion of the knowledge
distillation literature on the visual fidelity and closeness to original data,
we argue that the ultimate goal is not to accurately emulate the data-generating
distribution. Instead, it should be to sample effectively along the decision
boundary region, such that a student can later mimic it. With this in mind, we propose to create pairs of noisy synthetic samples and employ a
contrastive approach to diffuse them towards the decision boundary. Intuitively, think of drawing two samples for two different classes (or sets in multi-class scenarios) and pulling both towards each other until their predicted label gets swapped.
To this end, we employ the squared Euclidean distance
between logit pairs for synthetic samples of different classes:
\begin{equation}
	\label{eq:contr-loss}
	\fn{\mathcal{L}_{\text{contr}}}{\mbx_{i},\mbx_{j}} = \mathbb{1} \left[ y_{i} \neq y_{j} \right]\left\|\fn{\teachermodel}{\mbx_{i}} - \fn{\teachermodel}{\mbx_{j}}\right\|_{2}^{2} \quad .
\end{equation}
Note that despite the availability of elaborate contrastive formulations~\citep{oord2018InfoNCE,song2015lifted-structured-loss,frosst2019sonft-nn}, we focus on initial simplicity.
To avert the risk of synthetic samples collapsing into a single region
that minimizes the objective, recall the second row of \cref{fig:synth-examples}, it becomes necessary to further disperse these samples along the decision boundary, as we elaborate in the following subsection.

\subsection{Sweeping the Decision Boundary: Implicit and Explicit Noise}
\label{sec:cake:noise-and-schedules}

Having developed an objective aimed at generating samples close to the decision boundary, we must acknowledge that this objective
does not yet ensure extensive coverage \emph{along} the decision
boundary. On the one hand, abductive knowledge extraction need
not perfectly reflect the data distribution. However, on the other hand, it is imperative to mimic a wide range of the decision boundary. We therefore require an additional mechanism
to explore along the decision boundary. We posit that such exploration can be achieved through the introduction of noise into the sample update. As the contrastive term already acts as a perpendicular force, ensuring closeness between sets of samples between classes, the injection of noise effectively diffuses them in parallel to the decision boundary. In CAKE, we thus inject noise by means of the well-understood stochasticity of SGD-based optimizations and common step size schedules. Again for initial simplicity, we choose a simple linear schedule, but we note that a plethora of variants for noisy estimates exist. This effectively causes the optimization to disperse the synthetic samples along the decision boundary.

While CAKE presents an intuitive, highly empirically effective, but perhaps somewhat ad-hoc, solution to the induction of noise, we now also propose a more principled formulation. Recent advances
in generative modeling have rediscovered the importance of noise through the integration of
diffusion processes. Following this spirit, we introduce a CAKE variant termed Langevin Abductive Knowledge Extraction (LAKE). In the latter, we incorporate noise into the synthesis procedure with Langevin dynamics based diffusion, generating samples from noisy gradients of the input:
\begin{equation}
	\label{eq:langevin-step}
	\mbx_i^{t+1} = \mbx_i^{t} + \fn{\eta}{t} \nabla_{\boldsymbol{x}} \fn{\mathcal{L}}{\mbx_i^t} + \sqrt{2 \fn{\eta}{t}} \boldsymbol{\varepsilon}_i^{t} \quad ,
\end{equation}%
with $\boldsymbol{\varepsilon}_i^{t} \sim \mathcal{N}(0, \mathbf{I})$
for $t=1, \ldots , T$. The process will converge samples according to the true
distribution defined by the loss landscape, as both $T \rightarrow \infty$ and
$\fn{\eta}{t} \rightarrow 0$. The diffusion property of the Langevin update step
aids in dispersing samples along the decision boundary, thus preventing collapse. However, the theoretical guarantees only hold for the limit
$T \rightarrow \infty$ and $\fn{\eta}{t} \rightarrow 0$ and further empirical findings
seem to indicate that the explicit Gaussian noise term in the diffusion process
may not be fully necessary~\citep{bansal2022cold,daras2022soft}. Ultimately, we emphasize that the presence of noise seems to be crucial, as also highlighted by the quantitative results for CAKE and LAKE in subsequent \cref{sec:ablation-studies}, but the choice w.r.t a potential trade-off between empirical results and rigor is left to the prospective user.

\subsection{Injecting Auxiliary Domain-specific Knowledge: The Role of Data Priors}
\label{sec:cake:prior-knowledge}

In addition to our rigorous premise of no access to original training data, we acknowledge that information about the data domain typically exists. That is, even when a pre-trained model contains no reference to real data, its purpose and domain of application is typically made obvious. There is no conflict in integrating such auxiliary knowledge into the synthesis process through data priors. For instance, when the application is image-based, we can employ a total-variation prior, as initially used also by
DeepDream for the purpose of ``generating beautiful art'' from random noise:
\begin{equation}
	\label{eq:tv-loss}
	\fn{\mathcal{L}_{\text{TV}}}{\mbx} = \sum_{i=1}^{H}\sum_{j=1}^{W}\left\|\mbx_{i,j}-\mbx_{i-1,j}\right\|+\left\|\mbx_{i,j}-\mbx_{i,j-1}\right\| \, .
\end{equation}
Here, $\mbx$ represents an image of dimensions $H \times W$, and $\mbx_{i,j}$
corresponds to the pixel at the location $\left(i,j\right)$. Intuitively, this prior mirrors our expectation that inputs are images, and we thus expect depicted concepts to be locally consistent. More generally,  such priors enable injection of potential meta-knowledge we may possess in the form of constraints that facilitate the synthetic sample optimization. Whereas our work later showcases popular image classification, imagine e.g. a prior on the range of expected numerical values when confronted with tabular data as a second example.

\subsection{The Overall CAKE Algorithm}

For completeness, we lay out the full CAKE procedure in \cref{appendix:algorithm}. Conceptually, all synthetic samples could be generated in parallel. However, due to both practical compute and memory constraints, and to make the injection of noise more intuitive, the algorithm outlines the generation of $M$ sets of synthetic sample ``mini-batches''. For each mini-batch $\datasynth_m$, $N$ random synthetic samples and
labels $(\xsynth_{i}^{t=0}, \ysynth_{i})$, where $\xsynth_{i}^{t=0}$ and
$\ysynth_{i}$ are drawn from priors of our choice $\fn{p}{\mbx}, \fn{p}{y}$.
Subsequently, the algorithm iterates over the number of synthetic samples per
mini-batch, $\frac{N}{M}$, and for each sample, it performs $T$ iterations.
Within each iteration, the samples $\xsynth_{i}^{t}$ are fed through the teacher
$\teachermodel$ to obtain logits $\teacherlogits$. Then, we compute the
extraction loss $l$ as a weighted mixture of $\mathcal{L}_{\text{KD}}, \mathcal{L_{\text{contr}}}$, and $\mathcal{L}_{\text{TV}}$.
An update step with scheduled step size $\fn{\eta}{m}$ is performed on the
synthetic sample as specified in \cref{sec:cake:noise-and-schedules}. The
algorithm ultimately returns the union of all synthetic mini-batches,
$\datasynth = \bigcup_{m=1}^{M} \datasynth_{m}^T$. We can then proceed and train a
student model on the newly synthesized dataset. In CAKE, we argue that the necessary noise can be induced intuitively as a function of the current mini-batch.
Respectively, for LAKE we replace line 8 with earlier \cref{eq:langevin-step}.

\section{ABLATION STUDIES ON CIFAR}
\label{sec:ablation-studies}

To highlight the contributions of the design elements introduced in \cref{sec:cake:extraction-objectives,sec:cake:noise-and-schedules,sec:cake:prior-knowledge} and to corroborate their utility beyond the two-dimensional \cref{fig:synth-examples}, we start with ablation studies on CIFAR-10. \cref{tab:ablation-study} shows respectively obtained student accuracies for CAKE and the  LAKE variant for a ResNet-34~\citep{He2015DeepRL} teacher and smaller ResNet-18 student, both trained for 30 epochs on batches of size 256 with SGD and a learning rate of 0.5 scheduled with OneCycleLR~\citep{smith2018super}. We extract 500 mini-batches of 256 samples with loss weights $\lambda_{\text{contr}}=1\mathrm{e}1, \lambda_{\text{cls}}=1\mathrm{e}3$, and
$\lambda_{\text{TV}}=1\mathrm{e}5$ for 256 iterations and an initial step size
of 0.1. Further details follow standard practice and are provided in
\cref{appendix:algorithm} and \cref{appendix:hyperparams}. As described in earlier sections, we introduce noise by linearly decaying the step size across four magnitudes for the sample synthesis in CAKE and through Langevin diffusion in LAKE.

\begin{table}
	\caption{Ablation analysis of CAKE \& LAKE in distilling a ResNet-34 to a ResNet-18 on
		CIFAR-10, highlighting that inclusion of each individual component is meaningful to the
		overall performance of our method.}
	\label{tab:ablation-study}
	\begin{center}
		\begin{tabular}{l c c}
  \toprule
  & \multicolumn{2}{c}{Student Accuracy} \\
  \cmidrule(lr){2-3}
  Setting                                                                            & LAKE            & CAKE            \\
  \midrule
  baseline                                                                           & 28.0 $\pm$ 3.16 & 15.6 $\pm$ 4.15 \\
  $+ \mathcal{L}_{\text{KD}}$                                                        & 34.8 $\pm$ 5.56 & 19.2 $\pm$ 4.91 \\
  $+ \mathcal{L}_{\text{contr}}$                                                     & 24.7 $\pm$ 2.32 & 39.7 $\pm$ 7.27 \\
  $+ \mathcal{L}_{\text{KD}} + \mathcal{L}_{\text{contr}} $                          & 36.8 $\pm$ 3.02 & 42.5 $\pm$ 8.13 \\
  $+ \mathcal{L}_{\text{KD}} + \mathcal{L}_{\text{contr}} + \mathcal{L}_{\text{TV}}$ & 58.6 $\pm$ 4.02 & 71.0 $\pm$ 3.75 \\
  \bottomrule
\end{tabular}

	\end{center}
\end{table}
We can observe that the baseline, where synthetic samples are generated solely by
maximizing cross-entropy, delivers a student accuracy of only 28.0\% for LAKE
and 15.6\% for CAKE, demonstrating the necessity of additional
synthetization terms. The addition of the knowledge distillation loss
($\mathcal{L}_{\text{KD}}$) improves the performance for both LAKE and CAKE,
increasing student accuracy to 34.8\% and 19.2\% respectively, indicating, that
the use of teacher-based soft labels is not as effective when applied to
synthetic data compared to original training data. However, the
contrastive loss $\mathcal{L}_{\text{contr}}$ presents a more nuanced scenario.
Whereas it significantly enhances the performance of CAKE to 39.7\%, it seems to
slightly degrade performance for LAKE's to 24.7\% when taken on its own, suggesting that
$\mathcal{L}_{\text{contr}}$ is more beneficial in the absence of noise. However, when $\mathcal{L}_{\text{KD}}$ and
$\mathcal{L}_{\text{contr}}$ are combined, both LAKE and CAKE exhibit improved
performance, with student accuracy reaching 36.8\% and 42.5\% respectively,
illustrating the complementary nature of these two loss components. The final addition
of $\mathcal{L}_{\text{TV}}$ as a means to induce prior knowledge about the
possible structure of image-based data leads to large
improvements in performance of both LAKE and CAKE, resulting in student
accuracies of 58.6\% and 71.0\%.

\section{CAKE ACROSS SCALES}  %
\label{sec:model-scaling}
\begin{figure*}[h]
	\centering
	\input{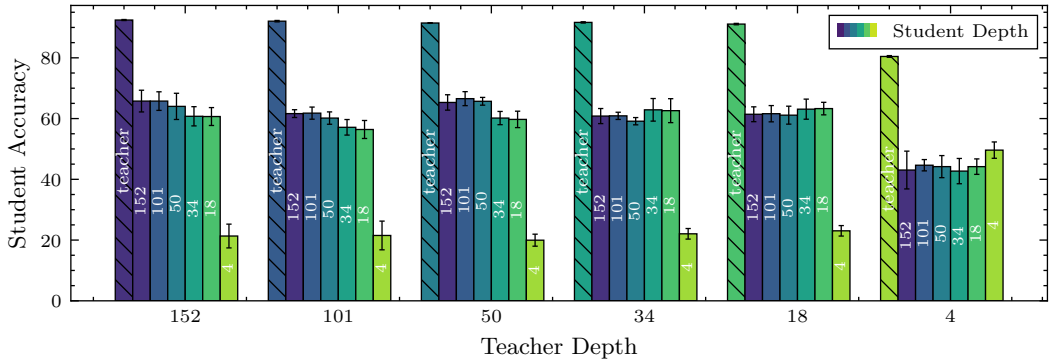}%
	\caption{Student model accuracy on CIFAR-10 (y-axis) when trained on synthetic data distilled from ResNet teacher models of different depths. Each group of bars corresponds to a ResNet teacher model of a particular depth (x-axis), and each bar within a group shows the accuracy of the student model distilled from that teacher model, along with its standard deviation as error bars. As desired, CAKE can compress models at a stable accuracy until capacity is too heavily constrained.}
	\label{fig:model-depth-distillation}
\end{figure*}

Following the spirit of the original knowledge distillation paper's experiments and goals~\citep{hinton2015distill}, we investigate CAKE's ability for model compression on CIFAR-10. To this end, we employ the well-known family of
ResNet models with varying depths as both teachers and students. Specifically, we consider the following depths, with the number of parameters denoted in millions in brackets: 152 (58.2M), 101 (43.5M), 50 (23.5M), 34 (21.3M), 18 (11.2M), and 4 (1.2M). \cref{fig:model-depth-distillation} shows respective groups of bars for achieved test accuracy for each teacher depth on the x-axis, where the hatched bars (with further explicit ``teacher'' bar label) show the respectively sized teacher's performance. All other bars quantify students of varying depths. For convenience, student depth is provided explicitly as the bar label and an applied shading further highlights smaller models through lighter shading.

From the obtained accuracies, it becomes evident that CAKE displays stable performance
across various teacher-student model capacities. Most importantly, even smaller
student models (ResNet-18, ResNet-34) exhibit competitive performance when
distilled from deeper teachers, indicating the effectiveness of CAKE in
compressing knowledge of previously overparametrized models.
As such, if the teacher model's complexity decreases too much, here in the case of ResNet-4, the student models also feature a significant drop in accuracy, irrespective of their depth. Naturally, this suggests that the lower capacity of the
teacher model limits the quality of knowledge it can provide, which the student cannot recover at any
capacity. Not surprisingly, a student model with very limited capacity suffers from bottlenecked information and amplifies performance degradation when knowledge is distilled, resulting in expected inferior
performance across all examined teacher model capacities.

\section{CAKE ACROSS MODEL TYPES}
\label{sec:different-teacher-and-student-models}

\begin{figure*}[h]
	\centering \input{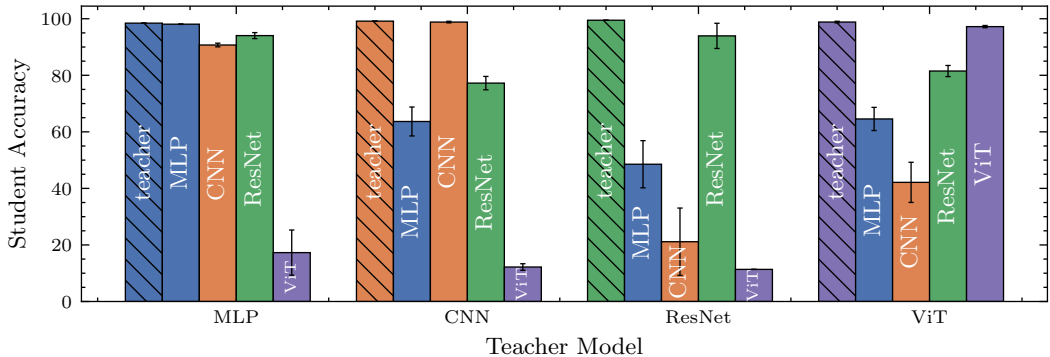}%
	\caption{Performance of different student models distilled from teacher models
		of various model types trained on MNIST: CNNs, MLPs, ResNets, and ViTs (parameter amounts are set to be similar). Each group of bars
		corresponds to a particular teacher type and each bar within a
		group shows the accuracy of a particular type of distilled student model, along with its standard deviation as
		error bars (5 trials). Overall, matching model types consistently provides good results, whereas distillation across types seems to work if the teacher has less inductive bias than the student.}
	\label{fig:model-cross-distillation}
\end{figure*}

A key advantage of CAKE does not only lie in its effectiveness without access to original data but also in the fact that there are no imposed constraints on model architectures or required intermediate model values for distillation. Much in contrast to the earlier mentioned prior works that require models to be of the same type or functioning on the premise of batch normalization layers, we are thus free to distill knowledge between a teacher and student model of different types. In fact, our sole requirement is that a model API implements a black box differentiable ``forward'' and a ``backward'' pass, where it is sufficient to simply obtain the final input gradient without any in-between states.

In the following, we thus investigate the performance of CAKE between four popular
neural network types: 1) \emph{Multi-layer Perceptrons (MLP)}, 2) \emph{Convolutional Neural Networks (CNN)}, 3)
\emph{ResNet-4}, and 4) \emph{Vision Transformer
	(ViT)}~\citep{dosovitskiy2021vit}. For fair comparison, we have matched the
models' parameter amounts, see appendix for details.
\cref{fig:model-cross-distillation} shows the result for \emph{across-model
	type} distillation on MNIST~\citep{lecun-mnisthandwrittendigit-2010}, combining every model type with every other. Each group on the x-axis describes a set of experiments with a specific teacher type, where the teacher's accuracy is hatched. The intra-group bars represent
student results when trained on the synthetic samples of the particular
teacher.

Overall, we find that distillation from an MLP to any other model is effective across the board, while the distillation performance from a ResNet to other models is generally
poor. Importantly, the distillation performance is notably robust when both the teacher and student
models share the same model type. Following these results, we first emphasize that our chosen models have all been roughly matched in terms of overall parameter amount and achieve negligibly similar teacher test accuracies. Thus, we hypothesize that when teacher and student models \emph{have similar inductive biases, the distillation process tends to be most
	effective}. In addition, as observable in the case of MLPs that have less inductive biases than the other contenders, it seems that the majority of students can also excel as they are unrestricted in forming their
own auxiliary assumptions. Here, the only exception is the ViT, for which distillation results are mixed and consistently underperform, unless the teacher is also a ViT.
We further conjecture that this outcome could be attributed to the fundamentally distinct manner in which inputs are fed into the model, specifically, the tokenization into sequences in ViTs.
However, most importantly, as we track \cref{fig:model-cross-distillation} to the right, our analysis thus suggests a rather simple rule of thumb. \emph{When in doubt of the teacher's type, choosing a ResNet model appears to be a safe choice}, as it provides stable performance independently of the source model.

\section{CAKE VS. TAILORED METHODS} %
\label{sec:comparison-other-methods}
\begin{table*}[h]
	\caption{Comparison of knowledge distillation techniques, presenting teacher
		and student model accuracies, to highlight that CAKE is effective despite
		lifting typical model constraints (MA: model-agnostic) and requiring no
		data access (DF: data-free). Note that standard deviations are typically
		not reported in the literature, obfuscating potential volatility in
		reproduction. See \cref{appendix:method-details} for further details on the compared methods.
	}
	\label{tab:comparison}
	\begin{center}
		\newcommand{\mr}[2]{\multirow{#1}{*}{#2}}
\newcommand{\Xmr}[1]{\multirow{#1}{*}{\xmark}}
\newcommand{\Cmr}[1]{\multirow{#1}{*}{\cmark}}
\begin{tabular}{l m{0.0075\textwidth} m{0.0075\textwidth} l l r l r}
  Method             & DF      & MA      & Dataset   & Teacher      & Acc.            & Student        & Acc.            \\
  \toprule
  \mr{3}{KD}         & \Xmr{3} & \Cmr{3} & MNIST     & LeNet-5      & 99.3            & LeNet-5-Half   & 98.8            \\% Taken from the ZSKD paper
                     &         &         & FMNIST    & LeNet-5      & 90.8            & LeNet-5-Half   & 89.7            \\% Taken from the ZSKD Paper
                     &         &         & CIFAR-10  & ResNet-34    & 95.6            & ResNet-18      & 94.3            \\% Taken from the DAFL paper
  \mr{2}{DAFL}       & \Cmr{2} & \Xmr{2} & MNIST     & LeNet-5      & 97.9            & LeNet-5-Half   & 97.6            \\% Taken from the RDSKD paper
                     &         &         & CIFAR-10  & ResNet-34    & 93.7            & ResNet18       & 90.4            \\% Taken from the RDSKD paper
    \mr{1}{DI}       & \Cmr{1} & \Xmr{1} & CIFAR-10  & ResNet-34    & 95.4            & ResNet-18      & 91.4            \\
  \mr{1}{ADI}        & \Cmr{1} & \Xmr{1} & CIFAR-10  & ResNet-34    & 95.4            & ResNet-18      & 93.3            \\
    \mr{1}{DD}       & \Cmr{1} & \Cmr{1} & CIFAR-10  & ResNet-34    & 95.4            & ResNet-18      & 30.0            \\% Taken from the DeepInversion paper
  \mr{3}{ZSDB3KD}    & \Cmr{3} & \Cmr{3} & MNIST     & LeNet-5      & 99.3            & LeNet-5-Half   & 96.5            \\
                     &         &         & FMNIST    & LeNet-5      & 91.6            & LeNet-5-Half   & 72.3            \\
                     &         &         & CIFAR-10  & AlexNet      & 79.3            & AlexNet-Half   & 59.5            \\
  \midrule
  \mr{7}{CAKE}       & \Cmr{7} & \Cmr{7} & MNIST     & LeNet-5      & 99.3 $\pm$ 0.12 & LeNet-5-Half   & 98.4 $\pm$ 0.18 \\
                     &         &         & FMNIST    & LeNet-5      & 91.0 $\pm$ 0.12 & LeNet-5-Half   & 76.5 $\pm$ 1.01 \\
                     &         &         & SVHN      & LeNet-5      & 89.8 $\pm$ 0.38 & LeNet-5-Half   & 62.9 $\pm$ 4.17 \\
                     &         &         & SVHN      & ViT-8        & 94.4 $\pm$ 0.13 & ViT-4          & 83.7 $\pm$ 4.77 \\
                     &         &         & SVHN      & ResNet-34    & 96.1 $\pm$ 0.08 & ResNet-18      & 94.2 $\pm$ 0.54 \\
                     &         &         & CIFAR-10  & ViT-8        & 73.2 $\pm$ 0.76 & ViT-4          & 53.8 $\pm$ 5.63 \\
                     &         &         & CIFAR-10  & ResNet-34    & 91.8 $\pm$ 0.11 & ResNet-18      & 78.9 $\pm$ 2.59 \\
  \bottomrule
\end{tabular}

	\end{center}
\end{table*}

Having evaluated the key factors contributing to CAKE's performance, its
efficacy in model compression, and its ability to distill across
diverse model types, we now position CAKE within the larger context of existing techniques. As discussed in
\cref{sec:knowledge-distillation}, these methods often require access to
original data, are tailored to specific models, or impose both
conditions. We include a wide set of techniques, their assumptions, and
performances on MNIST~\citep{lecun-mnisthandwrittendigit-2010},
FMNIST~\citep{xiao2017/online-fmnist}, SVHN~\citep{SVHN}, and
CIFAR~\citep{Krizhevsky2009LearningML} in \cref{tab:comparison} (see \cref{appendix:method-details}
for an extended table). Despite all other techniques imposing strict
requirements on model type and data availability, our results show compelling
evidence that CAKE can effectively lift assumptions with little to no
performance detriment.

For both MNIST (LeNet-5 to LeNet-5-Half) and SVHN (ResNet-34
to ResNet-18) settings, CAKE achieves comparable student accuracy to other techniques,
despite the latter requiring data access and/or being model-specific. In the
CIFAR-10 scenario (ResNet-34 to ResNet-18), we attain a student accuracy of
78.9\%, almost matching techniques with data access and additional model
assumptions with a mere 10\%-15\% gap. Remarkably, on CIFAR-10, CAKE outperforms
DeepDream (30.0\% for ResNet-34 to ResNet-18), the only other truly data-free
\textit{and} model-agnostic technique, by a factor of two. While ZSDB3KD is
data-free and model-agnostic as well, its focus is on model decision outputs and
is thus not directly comparable against DD and CAKE.

\section{DISCUSSION}
\label{sec:discussion-conclusion}

We have shown that CAKE can effectively transfer abductive knowledge between models of various capacities as well as models of entirely different types, despite the fact that CAKE lifts previous standard assumptions on models and requirements on original data access.
In light of these results, we challenge the current de facto standard of requiring original training data or making model assumptions. Already now, this entails a host of highly interesting future applications of societal significance, as well as an even greater set of prospects once remaining limitations are lifted.

\begin{figure}
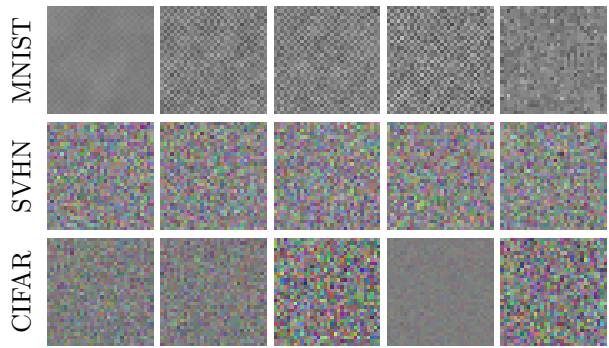

	\newcommand{\widthfactor}{1.0}
	\newcommand{\fig}[1]{\includegraphics[width=\widthfactor\linewidth]{ gfx/samples/#1-1x5-white.png }}
	\setlength{\tabcolsep}{2pt}
	\centering
	\begin{tabular}{m{0.04\linewidth} m{0.90\linewidth}}
		\rotatebox{90}{MNIST} & \fig{mnist}    \\
		\rotatebox{90}{SVHN}  & \fig{svhn}     \\
		\rotatebox{90}{CIFAR} & \fig{cifar-10} \\
	\end{tabular}
	\caption{Synthetic samples generated from a ResNet teacher by CAKE on the
		MNIST, SVHN, and CIFAR datasets, demonstrating no visual resemblance with original training data.}
	\label{fig:synth-samples-noisy}
\end{figure}

\paragraph{Future Work}
As CAKE's design lifts the requirement of original data access and
simultaneously removes unnecessary model constraints, it now opens up a plethora
of future applications and research directions. On the one hand, these lie in
performance improvements to our initial intuitive approach. For instance, we can
now further make use of the wide array of improved contrastive
formulations~\citep{oord2018InfoNCE,song2015lifted-structured-loss,frosst2019sonft-nn},
improvements for diffusion processes, or leverage adaptive signals similar to ADI~\citep{hongxu2019deepinversion} to
dynamically steer the distillation process based on the student's performance in
the spirit of curriculum learning~\citep{Wang2021ASO}. On the other, even more
exciting, hand, CAKE's main premise of extracting abductive knowledge also
entails that the data distribution is not closely mimicked. This implies that
generated synthetic samples do not resemble original data. \cref{fig:synth-samples-noisy}
showcases synthetic samples generated from a ResNet teacher using CAKE
across three datasets: MNIST, SVHN, and CIFAR. A crucial observation from the
depicted samples is their lack of visual resemblance to the original training
data from the respective datasets. Intuitively, they seem to look more like commonly found adversarial attacks~\citep{ilyas2019adv-bugs-not-feat}, but note that our synthetic data doesn't serve the same purpose to trick a classifier towards misclassification by perturbing original data.
This characteristic might offer a
foundational premise for robust privacy-preserving methodologies.
In light of these findings, future research could probe the potential of these
synthetic samples from a privacy perspective. One avenue to explore is the
application of differential privacy, which offer rigorous privacy guarantees by
ensuring that the release of statistical information doesn't compromise the
privacy of individual data entries. The intersection of synthetic data
generation and differential privacy could open up possibilities to ascertain:
(1) Whether these synthetic samples meet differential privacy criteria,
(2) the potential trade-offs between data utility and privacy when generating
such samples, and (3) the robustness of models trained on these synthetic
datasets against privacy attacks. In general, CAKE's ability to distill knowledge without
data resemblance could be invaluable for applications where data privacy and
confidentiality are paramount.

\paragraph{Limitations}
Although CAKE already yields promising results without common assumptions, we see two remaining limitations to be lifted in the future. First, as highlighted
in the previous paragraph, our current distillation process operates independently of the student model. This results in a lack of direct measures to estimate the quality of the synthetic dataset during the distillation phase, potentially limiting the effectiveness of the distillation and the resulting student model's performance. Second, although we assume no access to original data, model constraints, and require no access to intermediate model values, we do nevertheless still require a callable backward function. This does not yet allow CAKE to be used in scenarios where a model is hosted with an API that only allows forward evaluation calls, a limitation we foresee to be overcome through a transition to e.g. the very recent and concurrently proposed forward-forward algorithm~\citep{hinton2022forwardforward}.

\paragraph{Societal Impact}
We raise awareness that model-agnostic abductive knowledge extraction without training data access may be
misused to inappropriately extract knowledge from proprietary or confidential models, thereby leading to
potential violations of privacy or intellectual property theft. While amicable use cases for e.g. continual or federated learning exist, this also simultaneously highlights the importance of conducting further research into securing our public models. In particular, we foresee that the above-mentioned final limitation of requiring a backward API call may be lifted in the foreseeable future, exposing a crucial issue with current models. Finally, distillation methods will inadvertently mimic existing biases in the teacher model,
perpetuating or even exacerbating unfairness or discrimination, potentially making efforts towards data transparency even more challenging.

\section*{Acknowledgments}
This work was supported by the Federal Ministry of Education and Research (BMBF) Competence Center for AI and Labour (``kompAKI'', FKZ 02L19C150) and the project ``safeFBDC - Financial Big Data Cluster'' (FKZ: 01MK21002K), funded by the German Federal Ministry for Economics Affairs and Energy as part of the GAIA-x initiative. It benefited from the Hessian Ministry of Higher Education, Research, Science and the Arts (HMWK; projects ``The Third Wave of AI'' and ``The Adaptive Mind''), and the Hessian research priority programme LOEWE within the project ``WhiteBox''.

\bibliographystyle{plainnat}
\bibliography{bib}

\appendix
\onecolumn
\section{THE CAKE ALGORITHM}
\label{appendix:algorithm}
\begin{algorithm}
	\caption{Contrastive Abductive Knowledge Extraction}
	\label{algo:cake}
	\begin{algorithmic}[1]
		\Require teacher $\teachermodel$, iterations $T$, \#mini-batches
		$M$ of $N$ samples,
		step-size schedule $\eta$, priors $\fn{p}{\mbx}, \fn{p}{y}$
		\Procedure{CAKE}{$\teachermodel, T, M, N, \eta, \fn{p}{\mbx}, \fn{p}{y}$}
		\For{$m=1$ to $M$} \Comment{Number of mini-batches}
		\State Initialize $\datasynth_{m}^{t=0} \gets \left\{\left( \xsynth_{1}^{t=0}, \ysynth_{1}\right), \dots, \left(\xsynth_{N}^{t=0}, \ysynth_{N}\right)\right\}$, where $\xsynth_{i}^{t=0} \sim \fn{p}{\mbx}$ and $\ysynth_{i} \sim \fn{p}{y}$
		\For{$i=1$ to $N$} \Comment{Number of synthetic samples per mini-batch}
		\For{$t=1$ to $T$} \Comment{Number of iterations}
		\State $\teacherlogits \gets \fn{\teachermodel}{\xsynth_{i}^{t}}$ \Comment{Forward pass through teacher}
		\State $l \gets \fn{\mathcal{L}}{\xsynth_i^{t},\teacherlogits, \ysynth_{i}, \datasynth^t_m}$ \Comment{Compute extraction loss}
		\State
		$\xsynth_{i}^{t+1} \gets \xsynth_{i}^{t} - \fn{\eta}{m} \nabla_{\mbx} l$
		\Comment{Update synthetic samples}
		\EndFor
		\EndFor
		\EndFor
		\State \textbf{return} $\datasynth = \bigcup_{m=1}^{M} \datasynth_{m}^T$
		\EndProcedure
	\end{algorithmic}
\end{algorithm}

In \cref{algo:cake} we present the CAKE algorithm. The aim is
to extract knowledge from a given teacher model $\teachermodel$. The algorithm
requires a teacher model, the number of
iterations $T$, the number of mini-batches $M$ of $N$ samples, a learning rate
schedule $\eta$, and data and label priors $p(\mbx)$ and $p(y)$. The main procedure
starts with a loop generating $M$ mini-batches of size $N$ (Line 2). For every
mini-batch, synthetic samples are initialized (Line 3). These synthetic samples
$(\xsynth^{t=0}_i, \ysynth_i)$ are sampled from the priors with
$\xsynth^{t=0}_i \sim p(\mbx)$ and $\ysynth \sim p(y)$. Formally, the algorithm then optimizes each
of the $N$ sample within the mini-batch (Line 4). Implementation wise, this
for-loop can be implemented in a vectorized fashion by parallelizing the computations
over the mini-batch axis. For each sample,
an inner loop iterates for $T$ iterations (Line 5). In this loop, a forward
pass through the teacher model is conducted to compute $\teacherlogits$ (Line 6). Following
this, the objective loss $l$ is computed, contrasting the synthetic sample with
other samples in the batch and using the teacher's output (Line 7). The synthetic sample is then updated based on
this loss (Line 8). To implement LAKE as outlined in \cref{sec:cake:noise-and-schedules},
this line can be replace with the langevin dynamics update step from
\cref{eq:langevin-step}. Finally, after processing all samples and mini-batches,
we return the joint set of synthetic mini-batches $D$ (Line 9).

\section{CAKE HYPERPARAMETERS}
\label{appendix:hyperparams}
In our experiments, we defined a set of default hyperparameters for
CAKE that we used across all runs unless stated otherwise. This setup
involved extracting 500 mini-batches of 256 samples each, with the loss weights
$\lambda_{\text{contr}}=1\mathrm{e}1$, $\lambda_{\text{cls}}=1\mathrm{e}3$, and
$\lambda_{\text{TV}}=1\mathrm{e}5$ respectively. A mini-batch was generated over
256 update iterations with an initial step size of 0.1 and a linearly decaying
step schedule over four magnitudes. As priors in the sample generation process
we use a Gaussian distribution
$\xsynth_{i}^{t=0} \sim \fn{\mathcal{N}}{\mathbf{0}, \mathbf{I}}$ as data prior $\fn{p}{\mbx}$
and a uniformly distributed categorical distribution over the number of classes
$y_{i} \sim \fn{\text{Cat}}{\frac{1}{C}, \dots, \frac{1}{C}}$ as label prior $\fn{p}{y}$.
Furthermore, in the experiments discussed in
\cref{sec:model-scaling}, we reduced the number of mini-batches to 250 to
reduce the computational load. Conversely, for the experiments in
\cref{sec:comparison-other-methods}, the number of mini-batches was increased to
2000 to test the limits of CAKE.
One finding from our intermediate experiments was the noticeable improvement in
student accuracy as we increased the number of mini-batches. This observation
aligns with our intuitive understanding that a higher number of mini-batches
allows us to sample the data-relevant decision boundary regions more thoroughly
and in a fine-grained manner.

\section{NETWORK ARCHITECTURES, TRAINING, AND IMPLEMENTATION DETAILS}
\label{appendix:details}

\paragraph{Architectures.}
Below we outline the specific network architectures details utilized in our
experiments. For each model type, we provide information regarding their
architectural configurations, such as the number of layers and their dimensions,
as well as the total number of parameters involved. Each of these models has
been selected based on its relevance to the data sets used in our experiments,
and they represent a variety of model complexities and capabilities. Detailed
descriptions for each of these network architectures are provided below.

\begin{itemize}
	\item \textbf{Multi-layer Perceptron (MLP):} An architecture consisting of
	      four hidden layers, each having a hidden size of 100, amounting to a
	      total of 118K parameters.

	\item \textbf{Convolutional Neural Network (CNN):} The architecture of this
	      model featured four convolutional layers. The number of channels for
	      each layer was 32, 64, 64, and 64, respectively, with corresponding
	      kernel sizes of 3, 4, 3, and 3. We employed ReLU activation after each
	      convolution operation, and max pooling was introduced after the second,
	      third, and fourth convolutions. The architecture concluded with a fully
	      connected layer, totaling 107K parameters.

	\item \textbf{LeNet-5 \& LeNet-5-Half:} The LeNet-5 model adhered to the
	      architecture proposed by \citet{lecun1998lenet}, possessing 61.7K
	      parameters. In contrast, the LeNet-5-Half model was a modification of
	      LeNet-5, with the number of filters in each convolution layer reduced by
	      half, resulting in a compact model with 15.7K parameters.

	\item \textbf{Residual Network (ResNet):} The ResNet architectures with depths
	      of 152, 101, 50, 34, and 18 were utilized as described in
	      \citet{He2015DeepRL}. Furthermore, a modified ResNet with a depth of 4
	      was introduced, where the repetition factor of the layers \texttt{conv2}
	      to \texttt{conv5} was set to 1. The number of filters in these
	      convolution layers was configured based on the dataset, with 16, 32,
	      32, and 64 filters for MNIST and 32, 64, 128, and 256 filters for
	      CIFAR-10. The ResNet models accounted for 58M, 43M, 24M, 21M, 11M, 1.2M,
	      and 98K parameters correspondingly.

	\item \textbf{Vision Transformer (ViT):} We employed the ViT architecture for
	      small datasets as suggested by \citet{lee2021small-vit}. We had models
	      with depths of 8 (ViT-8) and 4 (ViT-4) with 8 heads, intermediate and
	      MLP dimensions of 64, and a patch size of 4, resulting in 1.1M and 580K
	      parameters respectively. For MNIST experiments, we further reduced model
	      capacity by using a model with a depth of 3 (ViT-3), 4 heads, and
	      intermediate and MLP dimensions of 32, resulting in a compact model with
	      110K parameters.
\end{itemize}

\paragraph{Training.}
In our experiments, both the teacher and student models were trained using the
following configurations. The training process was carried out over 30 epochs,
utilizing a batch size of 256 for each iteration. The models were optimized
using Stochastic Gradient Descent (SGD) with an initial learning rate of 0.5 and
weight decay of $1\mathrm{e-}4$. This learning rate was scheduled using the OneCycleLr~\cite{smith2018super}
learning rate scheduler with an initial division factor of 25 and a final
division factor of $1\mathrm{e}4$. For each experiment result we report the mean
and standard deviation over five runs with seeds $\left\{0, \dots, 4\right\}$.

\paragraph{Software and Hardware.}
All of our experiments were implemented in PyTorch (v2.0.0) for model
implementation and autograd and the Lightning framework (v2.0.0) for structuring
the training pipeline and facilitating model evaluation (see
\texttt{requirements.txt} for a full list of dependencies and their versions).
Experiments were run on A100 GPUs utilizing \texttt{bfloat16} precision.
The sample generation procedure on a ResNet-34 with a mini-batch size of 250 for
500 batches and 256 update steps per mini-batch takes 40 minutes with an average GPU utilization
of 96\% occupying 3.725GB of VRAM. All of our code is provided in the supplementary material for full reproducibility and
insights.

\section{DETAILS ON COMPARED METHODS}
\label{appendix:method-details}

\begin{table}[h]
	\caption{Comparison of knowledge distillation techniques, presenting teacher
		and student model accuracies, to highlight that CAKE is effective despite
		lifting typical model constraints (MA: model-agnostic) and requiring no data
		access (DF: data-free). Note that standard deviations are typically not
		reported in the literature, obfuscating potential volatility in
		reproduction.}
	\label{tab:full-comparison}
	\begin{center}
		\newcommand{\mr}[2]{\multirow{#1}{*}{#2}}
\newcommand{\Xmr}[1]{\multirow{#1}{*}{\xmark}}
\newcommand{\Cmr}[1]{\multirow{#1}{*}{\cmark}}
\begin{tabular}{l m{0.0075\textwidth} m{0.0075\textwidth} l l r l r}
  Method          & DF      & MA      & Dataset  & Teacher      & Acc.            & Student        & Acc.            \\
  \toprule
  \mr{3}{KD}      & \Xmr{3} & \Cmr{3} & MNIST    & LeNet-5      & 99.3            & LeNet-5-Half   & 98.8            \\% Taken from the ZSKD paper
                  &         &         & FMNIST   & LeNet-5      & 90.8            & LeNet-5-Half   & 89.7            \\% Taken from the ZSKD Paper
                  &         &         & CIFAR-10 & ResNet-34    & 95.6            & ResNet-18      & 94.3            \\% Taken from the DAFL paper
  \midrule
  \mr{4}{DAFL}    & \Cmr{4} & \Xmr{4} & MNIST    & LeNet-5      & 97.9            & LeNet-5-Half   & 97.6            \\% Taken from the RDSKD paper
                  &         &         & MNIST    & HintonNet    & 98.4            & HintonNet-Half & 97.9            \\
                  &         &         & SVHN     & WResNet-40-2 & 95.9            & WResNet-16-1   & 94.3            \\% Taken from the RDSKD paper
                  &         &         & CIFAR-10 & ResNet-34    & 93.7            & ResNet18       & 90.4            \\% Taken from the RDSKD paper
  \midrule
  \mr{3}{RDSKD}   & \Cmr{3} & \Xmr{3} & MNIST    & LeNet-5      & 97.9            & LeNet-5-Half   & 97.6            \\
                  &         &         & SVHN     & WResNet-40-2 & 95.9            & WResNet-16-1   & 94.6            \\
                  &         &         & CIFAR-10 & ResNet-34    & 93.7            & ResNet18       & 90.8            \\
  \midrule
  \mr{3}{DI}      & \Cmr{3} & \Xmr{3} & CIFAR-10 & VGG-11       & 92.3            & VGG-11         & 84.2            \\
                  &         &         & CIFAR-10 & VGG-11       & 92.3            & ResNet-18      & 83.8            \\
                  &         &         & CIFAR-10 & ResNet-34    & 95.4            & ResNet-18      & 91.4            \\
  \midrule
  \mr{3}{ADI}     & \Cmr{3} & \Xmr{3} & CIFAR-10 & VGG-11       & 92.3            & VGG-11         & 90.8            \\
                  &         &         & CIFAR-10 & VGG-11       & 92.3            & ResNet-18      & 90.7            \\
                  &         &         & CIFAR-10 & ResNet-34    & 95.4            & ResNet-18      & 93.3            \\
  \midrule
  \mr{5}{ZSKD}    & \Cmr{5} & \Xmr{5} & MNIST    & LeNet-5      & 97.9            & LeNet-5-Half   & 92.1            \\% Taken from the RDSKD Paper
                  &         &         & FMNIST   & LeNet-5      & 90.8            & LeNet-5-Half   & 79.6            \\
                  &         &         & SVHN     & WResNet-40-2 & 96.0            & WResNet-16-1   & 14.5            \\% Taken from the RDSKD Paper
                  &         &         & CIFAR-10 & AlexNet      & 83.0            & AlexNet-Half   & 69.6            \\% Taken from the RDSKD Paper
                  &         &         & CIFAR-10 & ResNet-34    & 93.7            & ResNet-18      & 10.5            \\% Taken from the ZSKD Paper
  \midrule
  \mr{3}{GD}      & \Cmr{3} & \Xmr{3} & SVHN     & ResNet-18    & 94.5            & MobileNetV2    & 92.9            \\
                  &         &         & CIFAR-10 & ResNet-34    & 93.3            & ResNet-18      & 86.0 $\pm$ 0.12 \\
                  &         &         & CIFAR-10 & ResNet-34    & 93.3            & ResNet-34      & 87.1 $\pm$ 0.23 \\
  \midrule
  \midrule
   \mr{3}{DD}     & \Cmr{3} & \Cmr{3} & CIFAR-10 & VGG-11       & 92.3            & VGG-11         & 36.6            \\% Taken from the DeepInversion paper
                  &         &         & CIFAR-10 & VGG-11       & 92.3            & ResNet-18      & 39.7            \\% Taken from the DeepInversion paper
                  &         &         & CIFAR-10 & ResNet-34    & 95.4            & ResNet-18      & 30.0            \\% Taken from the DeepInversion paper
  \midrule
  \mr{3}{ZSDB3KD} & \Cmr{3} & \Cmr{3} & MNIST    & LeNet-5      & 99.3            & LeNet-5-Half   & 96.5            \\
                  &         &         & FMNIST   & LeNet-5      & 91.6            & LeNet-5-Half   & 72.3            \\
                  &         &         & CIFAR-10 & AlexNet      & 79.3            & AlexNet-Half   & 59.5            \\
  \midrule
  \mr{7}{CAKE}    & \Cmr{7} & \Cmr{7} & MNIST    & LeNet-5      & 99.3 $\pm$ 0.12 & LeNet-5-Half   & 98.4 $\pm$ 0.18 \\
                  &         &         & FMNIST   & LeNet-5      & 91.0 $\pm$ 0.12 & LeNet-5-Half   & 76.4 $\pm$ 1.01 \\
                  &         &         & SVHN     & LeNet-5      & 89.8 $\pm$ 0.38 & LeNet-5-Half   & 62.9 $\pm$ 4.17 \\
                  &         &         & SVHN     & ViT-8        & 94.4 $\pm$ 0.13 & ViT-4          & 83.7 $\pm$ 4.77 \\
                  &         &         & SVHN     & ResNet-34    & 96.1 $\pm$ 0.08 & ResNet-18      & 94.2 $\pm$ 0.54 \\
                  &         &         & CIFAR-10 & ViT-8        & 73.2 $\pm$ 0.76 & ViT-4          & 53.8 $\pm$ 5.63 \\
                  &         &         & CIFAR-10 & ResNet-34    & 91.8 $\pm$ 0.11 & ResNet-18      & 78.9 $\pm$ 2.59 \\
  \bottomrule
\end{tabular}

	\end{center}
\end{table}

In this section, we will provide further information on the in
\cref{sec:comparison-other-methods} presented comparison against other relevant
KD methods.
We further extend \cref{tab:comparison} with more related work in \cref{tab:full-comparison}.
For each compared methods, we will reason on the distinctions we
made in \cref{tab:comparison}, in particular their attributes of being data-free
and model-agnostic.

\begin{itemize}
	\item \textbf{Knowledge Distillation} (KD)~\citep{hinton2015distill}: The
	      seminal work introduced concept of extracting ``soft targets'' from a
	      teacher model to train a student model. The results from
	      \cite{hinton2015distill} were obtained with access to the original
	      training data to fine-tune the student model, thereby inheriting the
	      data-dependency from its teacher model. Hence, it is not data-free. KD
	      has no further assumptions on the teacher or student model architecture
	      making it model-agnostic.

	\item \textbf{Data-Free Learning of Student Networks}
	      (DAFL)~\citep{chen2019dafl}: Utilizes Generative Adversarial Networks
	      (GANs) to train a compact student model without requiring access to the
	      original training data. The teacher network acts as a fixed
	      discriminator, and a generator produces synthetic samples for student
	      training, making the method data-free. However, DAFL incorporates an
	      architecture-specific activation loss term
	      \(\mathcal{L}_{a} = - \frac{1}{n} \sum_{i}||f_{T}^{i}||_{1}\) to guide
	      the learning. This loss term is designed to extract intrinsic patterns
	      from the teacher model's convolutional filters. Thus, DAFL's method is
	      not model-agnostic as it is (a) utilizes a GAN to synthesize samples and
	      (b) closely ties the GAN samples to the architecture of the
	      teacher model due to the architecture-specific activation loss.

	\item \textbf{Robustness and Diversity Seeking Data-free Knowledge
		      Distillation} (RDSKD)~\citep{han2020robustness-ad}:
	      This method addresses the limitations of existing data-free KD techniques by
	      incorporating a specialized loss function that focuses on sample authenticity,
	      class diversity, and inter-sample diversity. It mitigates the conflicting
	      goals of high sample authenticity and class diversity by exponentially
	      penalizing loss increments. Similar to DAFL, RDSK trains a generator to
	      synthesize samples

	\item \textbf{Zero-Shot Knowledge Distillation} (ZSKD)~\citep{nayak19azskd}:
	      This approach offers a data-free method for knowledge distillation, bypassing
	      the need for original training data or even meta-data. It synthesizes ``Data
	      Impressions'' from the teacher model, using them as surrogate data for
	      transferring learning to the student model. ZSKD constructs a class similarity
	      matrix and therefore constrains the model to have a fully-connected ultimate
	      layer with a soft-max non-linearity making it model-specific.

	\item \textbf{Gaussian Distillation}
	      (GD)~\citep{raikwar2022gaussian-distillation}: This technique employs samples
	      drawn from a Gaussian distribution for data-free knowledge distillation.
	      Unlike other attempts at using Gaussian noise, this method provides a reliable
	      solution by addressing the shift in the distribution of hidden layer
	      activations. GD addresses the challenge of ``covariate shift'' in hidden layer
	      activations through a modification in BatchNorm layers, making the method
	      depend on the presence of such layers in the teacher model.

	\item \textbf{DeepInversion} (DI)~\citep{hongxu2019deepinversion}: Generates
	      class-conditional input images by inverting a trained teacher network,
	      making it data-free. However, the method uses a feature distribution
	      regularization term \( \mathcal{R}_{\text{feature}} \) that depends on
	      BatchNorm layers in the teacher model to be present, rendering it not
	      model-agnostic.

	\item \textbf{Adversarial DeepInversion}
	      (ADI)~\citep{hongxu2019deepinversion}: An extension of DeepInversion.
	      ADI additionally maximizes the Jensen-Shannon divergence between teacher
	      and student network logits to improve synthesized image diversity.
	      Having the same constraints as DI, the adversarial extension now
	      directly creates a dependency between the teacher and student network.

	\item \textbf{DeepDream} (DD)~\citep{mordvintsev2015deepdream}: Originating as
	      a blog post, DeepDream serves as a technique for visualizing the learned
	      features of neural networks. It imposes no requirements on the original
	      training data or the architecture of the teacher model, making it both
	      data-free and model-agnostic. DeepDream creates enhanced images by
	      iteratively activating neurons in different layers of a pre-trained
	      network, giving us insights into what the network might "see" or
	      "imagine." This method is thus a fair comparator to CAKE, as it aligns
	      well with our criteria of being data-free and model-agnostic.

	\item \textbf{Zero-Shot Knowledge Distillation from a Decision-Based Black-Box
				Model} (ZSDB3KD)~\citep{pmlr-v139-wang21a}: Focuses on knowledge
				distillation when the teacher model is a decision-based black-box,
				providing only  hard labels (class predictions). To overcome this,
				ZSDB3KD uses the concept of ``sample robustness'' –- a measure of how
				far a sample lies from the teacher's decision boundaries. With
				accessible training data, sample robustness helps generate soft labels.
				When training data is absent (zero-shot), ZSDB3KD creates robust
				pseudo-samples mimicking the training data distribution. In both cases,
				knowledge distillation proceeds using generated soft labels.
				Importantly, ZSDB3KD requires neither the original training data nor
				specific details of the teacher's architecture, making it data-free and model-agnostic.\end{itemize}

\end{document}